%% file: main.tex
\documentclass[letterpaper, 10pt, journal, twoside]{IEEEtran}
\usepackage{amsmath,amsfonts}
\usepackage{algorithmic}
\usepackage{algorithm}
\usepackage{array}
\usepackage[caption=false,font=normalsize,labelfont=sf,textfont=sf]{subfig}
\usepackage{textcomp}
\usepackage{stfloats}
\usepackage{url}
\usepackage{verbatim}
\usepackage{graphicx}
\usepackage{siunitx}
\usepackage{balance}
\usepackage{cite}

\usepackage{xcolor} 
\usepackage{colortbl} 
\usepackage{bbding}
\usepackage{amssymb} 
\usepackage{booktabs} 
\usepackage{multirow, makecell}
\renewcommand{\arraystretch}{1.5} 
\usepackage{cleveref}
\crefname{section}{Sec.}{Secs.}
\crefname{figure}{Fig.}{Figs.}
\crefname{equation}{Eq.}{Eqs.}

\usepackage{placeins}   % \FloatBarrier
\usepackage{bm} 
\begin{document}

\title{Class-Distribution Guided Active Learning for 3D Occupancy Prediction in Autonomous Driving}

\author{Wonjune Kim$^{1}$, In-Jae Lee$^{2}$, Sihwan Hwang$^{3}$, Sanmin Kim$^{4\dagger}$, Dongsuk Kum$^{3\dagger}$
\thanks{Manuscript received: November, 18, 2025; Revised February, 15, 2026; Accepted March, 11, 2026. This paper was recommended for publication by Editor A. Banerjee upon evaluation of the Associate Editor and Reviewers' comments. This work was supported by Institute of Information \& communications Technology Planning \& Evaluation (IITP) and the National Research Foundation of Korea(NRF) funded by the Korea government(MSIT) under Grants RS-2023-00236245 and 2022R1A2C200494414.}
\thanks{$^{1}$Wonjune Kim is with the Robotics Program, KAIST, Daejeon 34141, South Korea, (email: {\tt\footnotesize wonjune.kim@kaist.ac.kr}).}
\thanks{$^{2}$In-Jae Lee is with the Interdisciplinary Program in Artificial Intelligence, Seoul National University, Seoul 08826, South Korea (email: {\tt\footnotesize injae.lee@snu.ac.kr}).}
\thanks{$^{3}$Sihwan Hwang and Dongsuk Kum are with the Graduate School of Mobility, KAIST, Daejeon 34051, South Korea (email: {\tt\footnotesize \{shhwang0129, dskum\}@kaist.ac.kr}).}
\thanks{$^{4}$Sanmin Kim is with the Department of Automobile and IT Convergence, Kookmin University, Seoul 02707, South Korea (email: {\tt\footnotesize sanmin.kim@kookmin.ac.kr}).}
\thanks{$^{\dagger}$Corresponding author: Sanmin Kim and Dongsuk Kum (email: sanmin.kim@kookmin.ac.kr; dskum@kaist.ac.kr).}%
\thanks{Digital Object Identifier (DOI): see top of this page.}
} % <-this % stops a space

\markboth{IEEE Robotics and Automation Letters. Preprint Version. Accepted March, 2026}
{Kim \MakeLowercase{\textit{et al.}}: Class-Distribution Guided Active Learning for 3D Occupancy Prediction in Autonomous Driving}

\maketitle

\input{class_color}
\input{sections/section_main}
\bibliographystyle{IEEEtran}
\bibliography{root}

\vfill

\end{document}

%% file: class_color.tex
\definecolor{barrier}{RGB}{112,128,144}
\definecolor{bicycle}{RGB}{220,20,60}
\definecolor{bus}{RGB}{255, 127, 80}
\definecolor{car}{RGB}{255, 158, 0}
\definecolor{const. veh.}{RGB}{233, 150, 70}
\definecolor{motorcycle}{RGB}{255,61,99}
\definecolor{pedestrian}{RGB}{0,0,230}
\definecolor{traffic cone}{RGB}{47,79,79}
\definecolor{trailer}{RGB}{255,140,0}
\definecolor{truck}{RGB}{255,99,71}
\definecolor{drive. suf.}{RGB}{0,207,191}
\definecolor{drive. surf.}{RGB}{0,207,191}
\definecolor{other flat}{RGB}{175,0,75}
\definecolor{sidewalk}{RGB}{75,0,75}
\definecolor{terrain}{RGB}{112,180,60}
\definecolor{manmade}{RGB}{222,184,135}
\definecolor{vegetation}{RGB}{0,175,0}
\definecolor{others}{RGB}{0, 0, 0}
\newcommand{\cmark}{\ding{51}}
\newcommand{\xmark}{\ding{55}}

%% file: sections/section_main.tex
\begin{abstract}
3D occupancy prediction provides dense spatial understanding critical for safe autonomous driving. However, this task suffers from a severe class imbalance due to its volumetric representation, where safety-critical objects (bicycles, traffic cones, pedestrians) occupy minimal voxels compared to dominant backgrounds. Additionally, voxel-level annotation is costly, yet dedicating effort to dominant classes is inefficient.
To address these challenges, we propose a class-distribution guided active learning framework for selecting training samples to annotate in autonomous driving datasets. Our approach combines three complementary criteria to select the training samples. Inter-sample diversity prioritizes samples whose predicted class distributions differ from those of the labeled set, intra-set diversity prevents redundant sampling within each acquisition cycle, and frequency-weighted uncertainty emphasizes rare classes by reweighting voxel-level entropy with inverse per-sample class proportions.
We ensure evaluation validity by using a geographically disjoint train/validation split of Occ3D-nuScenes, which reduces train–validation overlap and mitigates potential map memorization. With only 42.4\% labeled data, our framework reaches 26.62 mIoU, comparable to full supervision and outperforming active learning baselines at the same budget. We further validate generality on SemanticKITTI using a different
architecture, demonstrating consistent effectiveness across datasets.

\end{abstract}


%%%%%%%%%%%%%%%%%%%%%%%%%%%%%%%%%%%%%%%%%%%%%%%%%%%%%%%%%%%%%%%%%%%%%%%%%%%%%%%%%%%%%%%%%%%%%%%%%%%%%%%%%%%%%%%%%%%%%%%%%%%%%%%%%%%
\section{Introduction}
\IEEEPARstart{A}CCURATE autonomous driving perception enables essential tasks such as object detection, path planning, and collision avoidance, all of which contribute to the safety and efficiency of autonomous systems. However, conventional perception approaches face significant limitations in representing complex environmental structures. Traditional 3D object detection methods~\cite {yin2021center, liu2022bevfusion} typically represent objects using bounding boxes, which often fail to capture the complex geometries and irregularities of real-world objects, particularly in dynamic environments \cite{scene_as_occupancy, occ3d}. To overcome these limitations, 3D occupancy prediction has emerged as a promising alternative, offering a voxel-based representation of the environment that provides greater detail and coverage \cite{monoscene, zhang2023occformer, yu2023flashocc}.

\begin{figure}[t]
    \centering
    \includegraphics[width=0.49\textwidth]{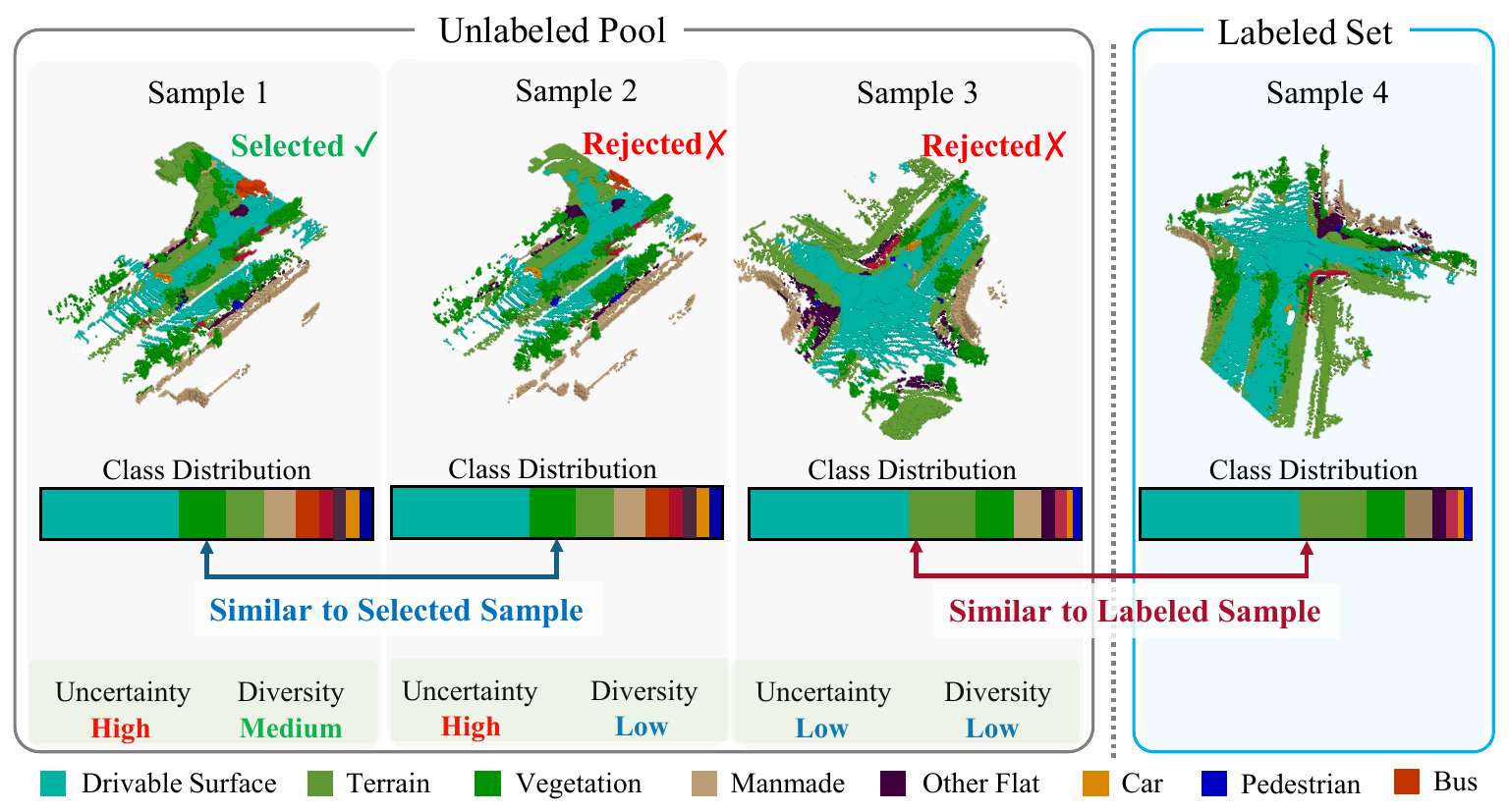}
    \caption{\textbf{Overview of the proposed class-distribution guided acquisition strategy based on predictive uncertainty and class-distribution diversity.}
Unlabeled samples receive an uncertainty score and a diversity score computed from similarity to the current selection and to the labeled set. The method selects samples that are uncertain and dissimilar to both sets and rejects redundant or certain samples. Sample~1 is selected because uncertainty is high and similarity to both sets is low. Sample~2 is rejected because it is similar to the selected Sample~1, which lowers intra-set diversity. Sample~3 is rejected because it is similar to the labeled reference Sample~4 and its uncertainty is low. Sample~4 is the labeled anchor used for similarity assessment. The bars show predicted class distributions.}
    \label{fig:intro_figure}
\end{figure}

Despite these advantages, transitioning to 3D occupancy prediction introduces two main challenges.  
First, voxel-wise annotation is extremely expensive. Each sample contains hundreds of thousands of voxels and every visible voxel requires a semantic label. Point-level LiDAR segmentation~\cite{behley2019semantickitti, zhu2021cylindrical} and box-level 3D object detection~\cite{yin2021center, liu2022bevfusion} only annotate observed surfaces or a small set of objects. In contrast, 3D occupancy prediction assigns labels to both occupied and free voxels within the visibility mask. As a result, labeling becomes slow and labor-intensive and the size of fully annotated occupancy datasets is effectively limited. 
Second, the volumetric representation induces a severe long-tail class distribution. Most voxels belong to free space and background classes such as drivable surfaces, roads, and buildings, whereas safety-critical objects such as pedestrians, bicycles, and traffic cones occupy only a tiny fraction of voxels. These minority objects appear in far fewer samples and cover far fewer voxels whenever they are present, which makes the effective sample size for rare classes extremely small. As a result, 3D occupancy grids are even more imbalanced than 3D object detection or point cloud segmentation tasks, and models trained on such skewed data tend to overfit frequent backgrounds while struggling to recognize rare categories that matter most for safety.

To address these two challenges, we adopt an active learning (AL) perspective and aim to allocate the annotation budget to samples that yield the greatest benefit~\cite{BADGE, hu2022lidal, wei2024basal, coreset}. However, most existing AL methods for 3D perception are designed for 3D object detection or point cloud segmentation, where only a small set of bounding boxes or surface points is labeled. Directly transferring these strategies to dense occupancy prediction is challenging, because occupancy requires voxel-wise labels for both occupied and free space, and uncertainty or diversity scores defined over sparse objects or points neither capture the distributional structure of voxel grids nor scale well to hundreds of thousands of voxels per sample. Object- or instance-level querying assumes sparse targets per sample, whereas 3D occupancy assigns labels to every voxel. Uncertainty-based strategies often focus on large ambiguous regions dominated by majority classes, which further reduces exposure to rare and safety-critical categories. Diversity-based strategies typically operate in a generic embedding space, providing no guarantee that minority semantics are adequately covered and incurring substantial memory and computation overhead when applied to high-resolution voxel grids~\cite{coreset, exploring_diversity_spatial, feng2019deep_mc-mi, hu2022lidal, wu2021redal, wei2024basal}. These limitations motivate a sample-level AL strategy that explicitly takes class-distribution structure into account and remains scalable in dense volumetric settings.

We propose an active learning framework that leverages class distributions to guide the selection of informative training samples for 3D occupancy prediction. \Cref{fig:intro_figure} illustrates how our acquisition strategy scores unlabeled samples and decides which ones to label. It favors samples whose predicted class distributions are both uncertain and dissimilar to those of the labeled set and the current selection, and rejects similar or low-uncertainty samples. Our selection policy uses three complementary scores.
\textbf{Inter-sample diversity} measures how different an unlabeled sample’s predicted class distribution is from those of labeled samples, encouraging novel class compositions (\Cref{sec:inter_diversity}). \textbf{Intra-set diversity} discourages near-duplicate samples within the same acquisition cycle (\Cref{sec:intra_diversity}). \textbf{Frequency-weighted uncertainty} aggregates voxel-wise entropy with inverse class-frequency weights so that uncertainty on rare classes has greater influence on the sample score (\Cref{sec:discriminative_entropy}).

We validate the proposed framework on Occ3D-nuScenes~\cite{occ3d} using a geographically disjoint train/validation split to mitigate overlap~\cite{lilja2024localization} and observe consistent gains over active learning baselines. We further evaluate on SemanticKITTI~\cite{behley2019semantickitti} with a different occupancy architecture, indicating that the benefit is not limited to one dataset or model. By combining inter-sample diversity, intra-set diversity, and frequency-weighted uncertainty, the approach improves overall accuracy and consistently boosts performance on small and minority classes such as bicycles, pedestrians, and traffic cones, enhancing the reliability of 3D occupancy prediction for autonomous driving.

To summarize, our main contributions are as follows:
\begin{itemize}
    \item We introduce a class-distribution guided active learning framework for 3D occupancy prediction that incorporates inter-sample diversity and intra-set diversity measures to select diverse and representative samples.
    \item  We propose a frequency-weighted uncertainty metric that amplifies uncertainty on rare and small classes by reweighting voxel-level entropy with inverse per-sample class proportions, thereby steering the acquisition process toward safety-critical categories.
    \item We show that the proposed method reaches near fully supervised performance using only a fraction of the Occ3D-nuScenes labels under geographically disjoint splits.
\end{itemize}

%%%%%%%%%%%%%%%%%%%%%%%%%%%%%%%%%%%%%%%%%%%%%%%%%%%%%%%%%%%%%%%%%%%%%%%%%%%%%%%%%%%%%%%%%%%%%%%%%%%%%%%%%%%%%%%%%%%%%%%%%%%%%%%%%
\section{Related Work}
\subsection{3D Occupancy Prediction in Autonomous Driving}
3D occupancy prediction estimates the occupancy state (occupied or free) and semantic category for each voxel or grid cell, providing a dense representation of the driving scene. Unlike object detection or instance segmentation, which focus on foreground objects, occupancy prediction seeks to model the entire environment, including background structures such as roads, buildings, and vegetation. % +model 추가.
In autonomous driving, benchmark 3D occupancy datasets such as Occ3D\cite{occ3d}, SurroundOcc\cite{surroundocc}, and OpenOccupancy\cite{openoccupancy} provide dense voxel annotations and visibility masks. These benchmarks enable comparable evaluation of occupancy models.
Despite these advancements, occupancy prediction still suffers from high voxel-level annotation cost and severe long-tail class imbalance.

To address these challenges, researchers have explored self-supervised~\cite{selfocc, liu2024let, gan2024gaussianocc} and semi-supervised learning strategies~\cite{li2025semi, pham2025semi} that reduce dependence on manual labels. These approaches do not decide which samples to annotate or explicitly counter the long tail, which motivates active learning as a mechanism to allocate a fixed labeling budget to samples with the highest expected informativeness.

\subsection{Active Learning for 3D Perception}
Active learning (AL) seeks to minimize annotation costs by intelligently selecting the most informative samples for labeling. The principal AL strategies are categorized into uncertainty-based and diversity-based approaches.

Uncertainty-driven AL methods prioritize samples where model predictions exhibit high entropy or variance, assuming that annotating uncertain instances leads to greater performance gains. Entropy-based selection focuses on frames with high bounding box entropy~\cite{crb}, while Bayesian methods such as Monte Carlo Dropout and deep ensembles estimate epistemic uncertainty~\cite{feng2019deep_mc-mi}. Some approaches also incorporate temporal uncertainty modeling, quantifying uncertainty by measuring prediction consistency across sequential frames~\cite{hu2022lidal}. However, uncertainty-based selection often exacerbates class imbalance, as high-entropy regions tend to correspond to majority-class voxels rather than rare categories.

Diversity-driven AL methods aim to maximize feature coverage by selecting samples that best represent the entire dataset distribution. Some methods cluster LiDAR embeddings to ensure diverse sample selection~\cite{shao2022active_spatial_structure, wu2021redal}. While diversity-driven selection improves data coverage, it can fail to address minority-class underrepresentation, as selection prioritizes overall feature diversity rather than rare-class instances.

3D object detection selects sparse boxes and point cloud segmentation labels only observed points, whereas dense 3D occupancy assigns semantics to every voxel. As a result, directly transferring active learning strategies from detection or segmentation is inadequate: query rules are defined at the object or point level, uncertainty heuristics are calibrated to surfaces rather than volumes, and embedding-based diversity scales poorly to high-resolution voxel grids. Many prior pipelines also assume LiDAR-only inputs and object-level supervision, which misaligns with voxel-wise prediction. To address these gaps, our method adopts class-distribution aware querying that increases exposure to rare categories while preserving data coverage and diversity.

%%%%%%%%%%%%%%%%%%%%%%%%%%%%%%%%%%%%%%%%%%%%%%%%%%%%%%%%%%%%%%%%%%%%%%%%%%%%%%%%%%%%%%%%%%%%%%%%%%%%%%%%%%%%%%%%%%%%%%%%%%%%%%%%%%%

\section{Methodology}
\begin{figure*}[t!]
    \centering
    \includegraphics[width=\linewidth]{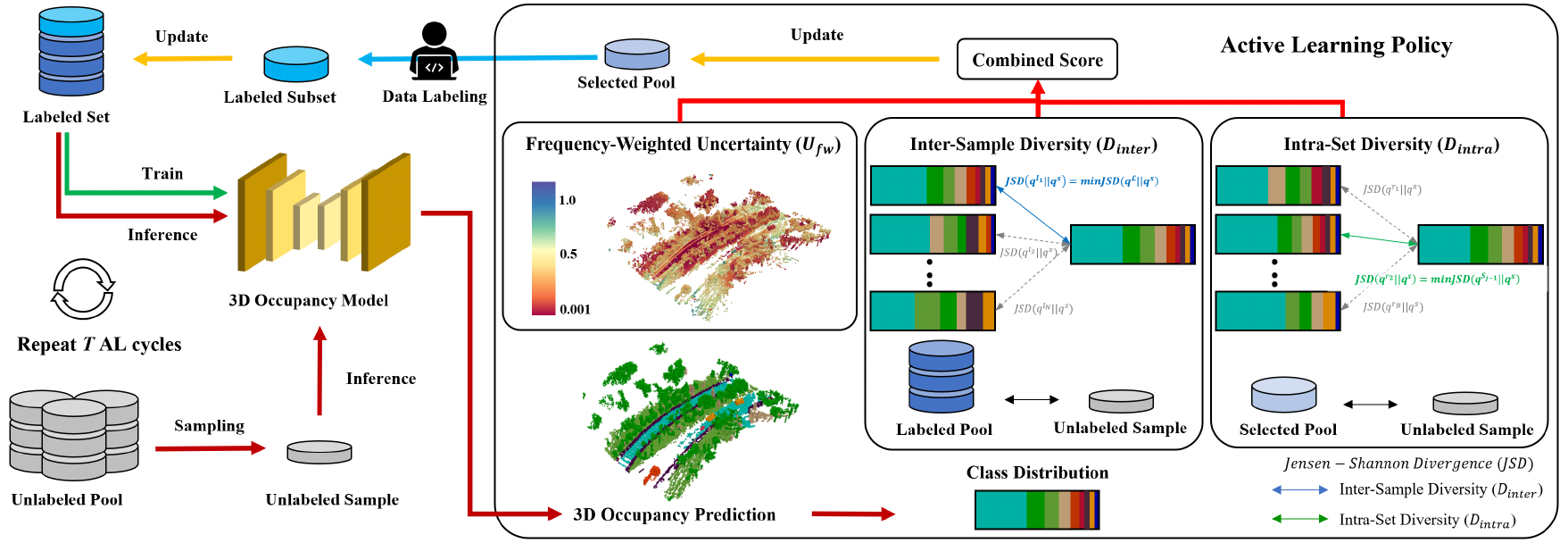}
    \caption{\textbf{Overview of our active learning method for 3D occupancy prediction.} An initial labeled set trains the model, which then performs inference on the unlabeled pool. In each iteration, candidate samples are scored using three complementary metrics: frequency-weighted uncertainty, inter-sample diversity relative to the labeled set, and intra-set diversity within the current selected samples. Scores are normalized and combined into a unified acquisition score. Top-ranked samples are selected for annotation and added to the labeled subset. The process repeats for $T$ cycles, each time updating the labeled set and retraining the model.}
    \label{fig:main_figure}
\end{figure*}

As shown in \Cref{fig:main_figure}, we propose a class-distribution guided active learning strategy that prioritizes samples that are both informative and diverse in their class distributions. Our framework operates in iterative cycles to efficiently label the most valuable 3D occupancy samples. We begin with an initial labeled set to train a 3D occupancy prediction model and then run inference on both labeled and unlabeled pools to obtain voxel-wise probabilities and sample-level class distributions. For each unlabeled sample we compute three complementary criteria that capture different aspects of usefulness: an uncertainty score that amplifies rare classes, an inter-sample diversity term between labeled set and unlabeled samples, and an intra-set diversity term within the unlabeled pool. These criteria are combined into a single acquisition score for each candidate sample. After scoring the unlabeled pool, we select the top-ranked samples for annotation, append them to the labeled set, and retrain the model on the expanded training set.

\subsection{Inter-Sample Diversity}
\label{sec:inter_diversity}

Uncertainty sampling alone does not prevent redundancy because high-uncertainty samples often resemble class distributions already abundant in the labeled set. For example, the model might be uncertain about certain regions in multiple highway samples that all have similar class compositions. If we keep selecting such similar samples, we miss the opportunity to label underrepresented compositions. To encourage distributional dissimilarity in selection, we introduce an inter-sample diversity criterion that favors samples with class distributions not present in the current training set. For each sample $s$ with $N$ voxels and $K$ classes, the model produces voxel-level class probabilities $p_i(c)$ at voxel $i$ for class $c\in\{1,\dots,K\}$.
Here, $N$ is the number of voxels within the visibility mask.
We compute the sample-level predicted class distribution as the proportion of voxels assigned to each class,
\begin{equation}
q_c \;=\; \frac{1}{N}\sum_{i=1}^{N} \mathbf{1}\!\left[\arg\max_{k} p_i(k) = c\right],
\qquad \sum_{c=1}^{K} q_c = 1.
\label{eq:scene_dist}
\end{equation}
We represent sample $s$ by the vector $q^{(s)}=\big(q^{(s)}_1,\dots,q^{(s)}_K\big)\in[0,1]^K$ computed by Eq.~\eqref{eq:scene_dist}, and let $\mathcal{L}$ denote the labeled set with sample-wise distributions $q^{(\ell)}$ for $\ell\in\mathcal{L}$. To quantify how different an unlabeled sample is from all previously labeled samples, we use the symmetric and bounded Jensen--Shannon divergence (JSD)~\cite{fuglede2004jensen}.
We compute $\mathrm{JSD}(\cdot\|\cdot)$ with $\log_2$, so $\mathrm{JSD}\in[0,1]$.
We then define the inter-sample diversity as the minimum divergence to the labeled set,
\begin{equation}
D_{\mathrm{inter}}(s) \;=\; \min_{\ell\in\mathcal{L}} \mathrm{JSD}\!\big(q^{(s)} \,\Vert\, q^{(\ell)}\big).
\label{eq:isd_minjsd}
\end{equation}
A larger $D_{\mathrm{inter}}(s)$ indicates that no labeled sample has a similar class composition, which promotes underrepresented distributions and reduces redundancy in the selected set.
% ------------------------------------------------------------
% ------------------------------------------------------------
\subsection{Intra-Set Diversity}
\label{sec:intra_diversity}

In each active learning cycle, we sequentially select $M$ samples. Let $j \in \{1,\dots,M\}$ denote the index of the current selection within the cycle, and let $\mathcal{S}_{j-1}$ be the set of samples that have already been chosen after $j-1$ selections. By definition, the selection set is empty at the beginning of the cycle, so $\mathcal{S}_{0} = \varnothing$.

Without additional constraints, a set of top-scoring samples can still contain several near-duplicates. Selecting redundant samples within the same cycle wastes annotation effort because labeling one of them makes the others largely redundant. To prevent this inefficiency, we introduce an intra-set diversity ($D_{\mathrm{intra}}$) criterion that explicitly promotes diversity across the samples chosen in the current cycle.

We measure similarity between samples via their predicted class distributions. For any sample $s$, the sample-level distribution $q^{(s)}$ is computed exactly as in Eq.~\eqref{eq:scene_dist}. For $j \ge 2$, we define the intra-set diversity score of a candidate $s$ conditioned on the current selection set as the nearest-neighbor divergence
\begin{equation}
\begin{aligned}
D_{\mathrm{intra}}\big(s \mid \mathcal{S}_{j-1}\big)
&= \min_{r \in \mathcal{S}_{j-1}} \mathrm{JSD}\!\big(q^{(s)} \,\Vert\, q^{(r)}\big), \quad j \ge 2.
\end{aligned}
\label{eq:ssd_minjsd}
\end{equation}
Since $\mathrm{JSD}\in[0,1]$, higher $D_{\mathrm{intra}}(s \mid \mathcal{S}_{j-1})$ means the candidate is more dissimilar to all samples already selected in the cycle.

Operationally, the $D_{\mathrm{intra}}$ term downweights candidates whose predicted class histograms are already well represented in $\mathcal{S}_{j-1}$.
For the first selection in each cycle ($j=1$), the set $\mathcal{S}_{0}$ is empty and $D_{\mathrm{intra}}$ is not used. The first sample is chosen using only the inter-sample diversity term and the uncertainty term that appears in our acquisition score (see Sec.~\ref{sec:aggregation}). After each new selection $s_j^\star$ with $j \ge 2$, the set is updated as $\mathcal{S}_{j}=\mathcal{S}_{j-1}\cup\{s_j^\star\}$ and the scores in Eq.~\eqref{eq:ssd_minjsd} are recomputed with only a single new divergence per remaining candidate, which keeps the procedure efficient.

\subsection{Frequency-Weighted Uncertainty}
\label{sec:discriminative_entropy}

Standard uncertainty-based sampling tends to prioritize regions with many uncertain predictions, which in 3D occupancy often correspond to large, majority-class areas such as roads or vegetation. As a result, traditional entropy criteria can overlook small or rare-class objects (e.g., traffic cones, bicycles) because these minority classes occupy only a very small fraction of voxels, often orders of magnitude fewer than background classes. Consequently, the model's uncertainty on rare classes contributes negligibly to a naive entropy score and is largely ignored, leaving those rare but important regions underrepresented in the selected data.

Given the per-voxel class probabilities $p_i(c)$ for class $c \in \{1,\dots,K\}$ at voxel $i$, the standard entropy at each voxel is defined as
\begin{equation}
H_i \;=\; -\sum_{c=1}^{K} p_i(c)\,\log p_i(c).
\label{eq:voxel_entropy}
\end{equation}
A conventional entropy-based sample score aggregates these voxel entropies uniformly over the grid and is therefore dominated by uncertainty on frequent background classes.

To address this bias, we introduce a frequency-weighted uncertainty metric. Using the sample-level class distribution $q_c$ defined in Eq.~\eqref{eq:scene_dist}, we weight each class's contribution to the sample's uncertainty by the inverse of its predicted prevalence in that sample. Concretely, we form inverse-prevalence weights
\begin{equation}
\tilde w_c \;=\; \frac{1}{q_c + \varepsilon}, 
\qquad
w_c \;=\; \frac{\tilde w_c}{\sum_{k=1}^{K} \tilde w_k},
\label{eq:invprev_weights_raw}
\end{equation}
with a small constant $\varepsilon=10^{-6}$ for numerical stability. These weights amplify rare classes while down-weighting dominant ones. The resulting class-weighted voxel entropy is
\begin{equation}
\widetilde H_i \;=\; -\sum_{c=1}^{K} w_c\, p_i(c)\,\log p_i(c),
\label{eq:weighted_entropy_voxel}
\end{equation}
and averaging over voxels yields the sample-level frequency-weighted uncertainty
\begin{equation}
U_{\mathrm{fw}}(s)=\frac{1}{N}\sum_{i=1}^{N}\tilde{H}_{i}
\;=\; -\frac{1}{N}\sum_{i=1}^{N}\sum_{c=1}^{K} w_c\, p_i(c)\,\log p_i(c).
\label{eq:U_rc}
\end{equation}

This score highlights samples where the model is particularly unsure about minor classes. A high uncertainty indicates that the sample contains rare-class content that the model finds confusing. Such samples are prioritized for labeling, since they promise to inform the model about classes it currently struggles with. In this way, the uncertainty criterion explicitly drives the selection of samples rich in minority-class information and helps to mitigate the class imbalance in the training data.

% ------------------------------------------------------------
% ------------------------------------------------------------

\subsection{Combined Acquisition Score}
\label{sec:aggregation}

Each unlabeled sample $s$ in the current pool is summarized by three complementary metrics:
\begin{itemize}
    \item \emph{Inter-Sample Diversity}: $D_{\mathrm{inter}}(s)$ (\Cref{sec:inter_diversity}),
    \item \emph{Intra-Set Diversity}: $D_{\mathrm{intra}}\!\big(s \mid \mathcal{S}_{j-1}\big)$ (\Cref{sec:intra_diversity}),
    \item \emph{Frequency-Weighted Uncertainty}: $U_{\mathrm{fw}}(s)$ (\Cref{sec:discriminative_entropy}).
\end{itemize}

To place these metrics on comparable scales, we apply the same robust normalization within the candidate pool: for each metric we subtract the median, divide by the interquartile range (IQR), and then apply min–max scaling to map values to $[0,1]$. This yields normalized scores
$\tilde D_{\mathrm{inter}}(s)$,
$\tilde D_{\mathrm{intra}}\!\big(s \mid \mathcal{S}_{j-1}\big)$,
and $\tilde U_{\mathrm{fw}}(s)$.

We then define the Combined Acquisition Score (CAS) at selection step $j$ as
\begin{equation}
\resizebox{0.9\linewidth}{!}{$
\mathrm{CAS}(s \mid \mathcal{S}_{j-1})
= \big\| \big(
    \tilde D_{\mathrm{inter}}(s),\,
    \tilde D_{\mathrm{intra}}(s \mid \mathcal{S}_{j-1}),\,
    \tilde U_{\mathrm{fw}}(s)
  \big) \big\|_{2}.
$}
\label{eq:cas_full}
\end{equation}
For the first selection in a cycle ($j=1$), the set $\mathcal{S}_{0}$ is empty and the intra-set term is not defined. By convention we set $\tilde D_{\mathrm{intra}}\!\big(s \mid \mathcal{S}_{0}\big)=0$ for all candidates, so the first sample is chosen using only inter-sample diversity and frequency-weighted uncertainty. For $j \ge 2$, all three normalized terms contribute to the score.

This $\ell_2$ aggregation is hyperparameter-free and balances uncertainty, novelty with respect to the labeled set, and within-batch diversity. We denote the current unlabeled candidate pool by $\mathcal{U}$. At step $j$, we greedily select
$s_j^\star \in \mathcal{U}\setminus\mathcal{S}_{j-1}$ as the candidate that maximizes
$\mathrm{CAS}(s \mid \mathcal{S}_{j-1})$.
We then update $\mathcal{S}_{j}=\mathcal{S}_{j-1}\cup\{s_j^\star\}$ and recompute only the intra-set diversity scores for the remaining candidates.

 In practice, we cache pairwise Jensen--Shannon divergences between sample-level class distributions to speed up updates within each acquisition cycle. For million-scale unlabeled pools, we avoid storing the full $O(|\mathcal{U}|^2)$ JSD matrix by using FAISS~\cite{johnson2019billion} for approximate nearest neighbor retrieval in the Hellinger embedding ($z=\sqrt{q}$), and compute exact JSD only over the retrieved candidates.

%%%%%%%%%%%%%%%%%%%%%%%%%%%%%%%%%%%%%%%%%%%%%%%%%%%%%%%%%%%%%%%%%%%%%%%%%%%%%%%%%%%%%%%%%%%%%%%%%%%%%%%%%%%%%%%%%%%%%%%%%%%%%%%%%%%
\input{tables/main_results_geo_split}
\input{tables/main_results_semkitti_short}
\section{Experiments}
\textbf{Implementation Details}: We adopt FlashOCC~\cite{yu2023flashocc} as the baseline architecture for Occ3D-nuScenes~\cite{occ3d} due to its simple yet competitive design. Occ3D-nuScenes defines 18 classes, including a free-space class. We report mIoU over the 17 semantic classes, excluding free-space. Since the official Occ3D-nuScenes test labels are not publicly available, we evaluate only on the public data and report results on a geographically disjoint train/validation split. To mitigate train--validation leakage from geographical overlap, we follow the geographically disjoint protocol of Lilja et al.~\cite{lilja2024localization} by removing boundary-crossing sequences and excluding sequences from the original test split. The resulting split contains 23,574 training samples and 5,118 validation samples. We run active learning for five cycles, adding a fixed budget of the training set at each cycle and retraining after every acquisition (\Cref{table:main_results_geo_split}). To validate generality beyond Occ3D and a single architecture, we additionally evaluate on SemanticKITTI~\cite{behley2019semantickitti} using OccFormer~\cite{zhang2023occformer}. We follow the official split (train: 00--07, 09, 10; val: 08) and report results exclusively on validation, since the test set is restricted to benchmark submissions and is not suitable for multi-cycle active learning (\Cref{table:main_results_semkitti_short}).

\subsection{Main Results}

\paragraph{Comprehensive Performance Analysis}
\Cref{table:main_results_geo_split} reports mIoU across active learning cycles on the geographically disjoint train/validation split~\cite{lilja2024localization} of Occ3D-nuScenes~\cite{occ3d}. At Cycle 5, using only 42.4\% of the training data, our method achieves 26.62 mIoU, comparable to the fully supervised result on this split (26.83 mIoU). This demonstrates significant label efficiency while maintaining near-optimal accuracy. The approach jointly leverages frequency-weighted uncertainty and sample-level class distributions so that the selected samples are both informative and diverse. This design raises overall mIoU already in early cycles and yields improvements on safety-critical foreground objects (\Cref{table:main_results_geo_split}).
We further validate this behavior on SemanticKITTI~\cite{behley2019semantickitti} using a different backbone (OccFormer~\cite{zhang2023occformer}), where our method remains consistently effective across cycles (\Cref{table:main_results_semkitti_short}).

Quantitatively, Coreset and Entropy are less effective than our method for different reasons. Coreset relies on embedding-based diversity that was developed for sparse object or point representations, and in dense voxel occupancy this limits its ability to intentionally increase exposure to rare foreground categories. On the other hand, Entropy is class-agnostic and tends to prioritize high-uncertainty regions that often occur in background or free space, which can divert labeling budget away from informative foreground samples. For some background categories, the gap to other methods is small or even slightly negative (\Cref{table:main_results_geo_split}). This reflects that leveraging frequency-weighted uncertainty with inter-sample and intra-set diversity leads more labels to foreground content with a smaller voxel proportion, reducing the variety of background patterns collected. Background categories dominate the dataset and vary less, so their performance saturates quickly, whereas missed foreground objects have a more direct impact on motion planning and collision avoidance. From a safety perspective, it is therefore more important to accurately detect foreground classes than to avoid small decreases in background-class performance.

\begin{figure*}[t!]
    \centering
    \includegraphics[width=\textwidth, clip]{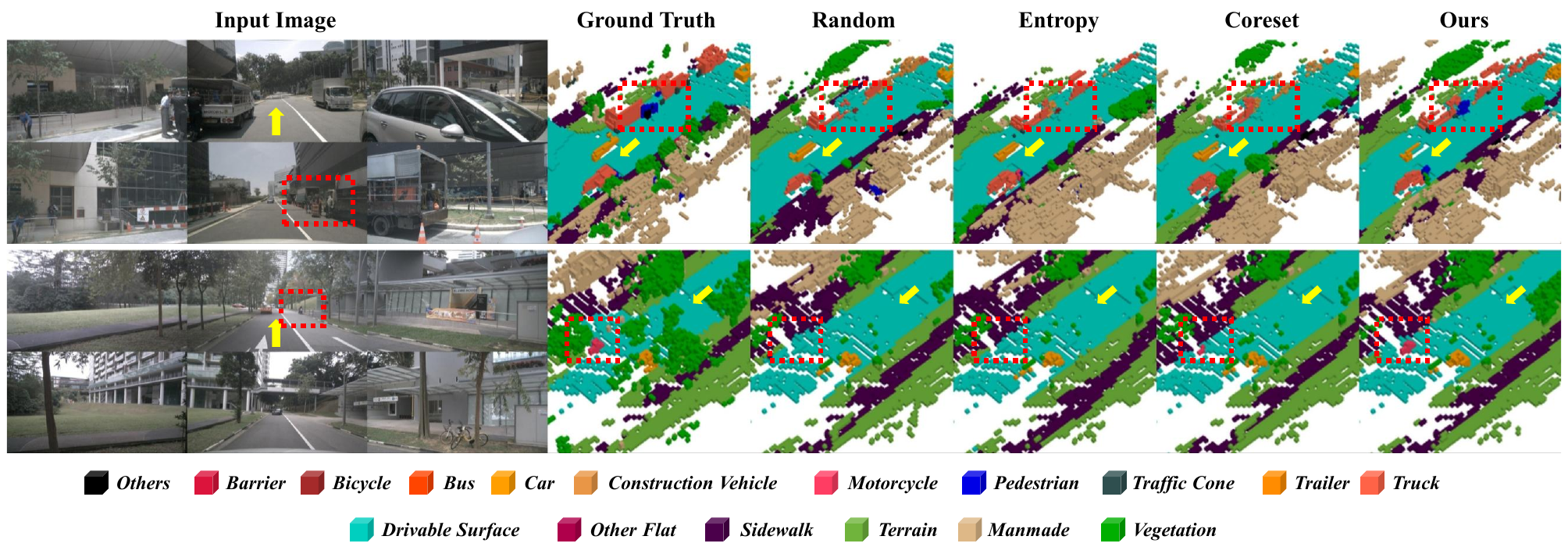}
    
    \caption{\textbf{Qualitative comparison of active learning strategies for 3D occupancy prediction.} Each example shows six surround-view images and the ground truth occupancy grid, followed by model outputs when samples are chosen by random, entropy, Coreset\cite{coreset}, and our active learning methods. Red dashed boxes mark regions where our approach more accurately predicts occupancy than the baselines. This improvement comes from our sample selection strategy, which emphasizes rare-class uncertainty and diversity in sample-level class distributions.}
    \label{fig:qualitative_results}
\end{figure*}

Qualitative results in \Cref{fig:qualitative_results} show fewer missed detections of pedestrians and bicycles compared with other methods and fewer cases where these objects are predicted as free space or assigned only a negligible number of voxels to other classes, which leads to more complete foreground occupancy.

\input{tables/night_rainy_results_geo_split}

\paragraph{Performance under night and rainy scenarios}
\Cref{tab:night_rainy_results_geo_split} reports cycle-wise mIoU on the night and rainy subsets, where each cell shows mIoU (top) and the number of selected samples (bottom).

For night samples, the entropy-based method tends to perform better across cycles, while our method is slightly less competitive.
One contributing factor is the acquisition distribution.
Our strategy selects substantially fewer night samples than entropy (e.g., Cycle~2: 59 vs.\ 795), which results in less night-time supervision during training.
This behavior is closely related to prediction reliability under low illumination.
When visibility is degraded, voxel predictions become broadly uncertain and class boundaries are blurred, which makes the sample-level class distribution estimates and frequency-weighted uncertainty noisier.
Entropy, on the other hand, assigns higher scores to these low-visibility samples and therefore acquires many more night samples.
The larger amount of labeled night data then supports better night-time performance.
One simple remedy is to add a lightweight visibility-aware adjustment during acquisition (e.g., an image-brightness proxy) to avoid systematically under-sampling low-visibility night samples.

For rainy samples, our method performs best or near-best across cycles and reaches 27.79 mIoU at Cycle~5.
This gain is not explained by selecting more rainy samples.
This trend is expected because most rainy sequences are captured in daylight, so voxel-level predictions remain sufficiently confident for reliable class estimates.
Moreover, light-to-moderate precipitation typically does not severely disrupt large-scale geometry (e.g., roads and buildings) nor the visibility of salient foreground instances (e.g., cars and pedestrians).
As a result, object extents and boundaries remain discernible, and the resulting sample-level class distributions are more reliable.
Under these conditions, our distribution-guided acquisition operates as intended and yields consistent improvements across cycles.

\begin{figure*}[t!]
    \centering
    \setlength{\abovecaptionskip}{3pt}
    \includegraphics[width=1.0\textwidth, clip]{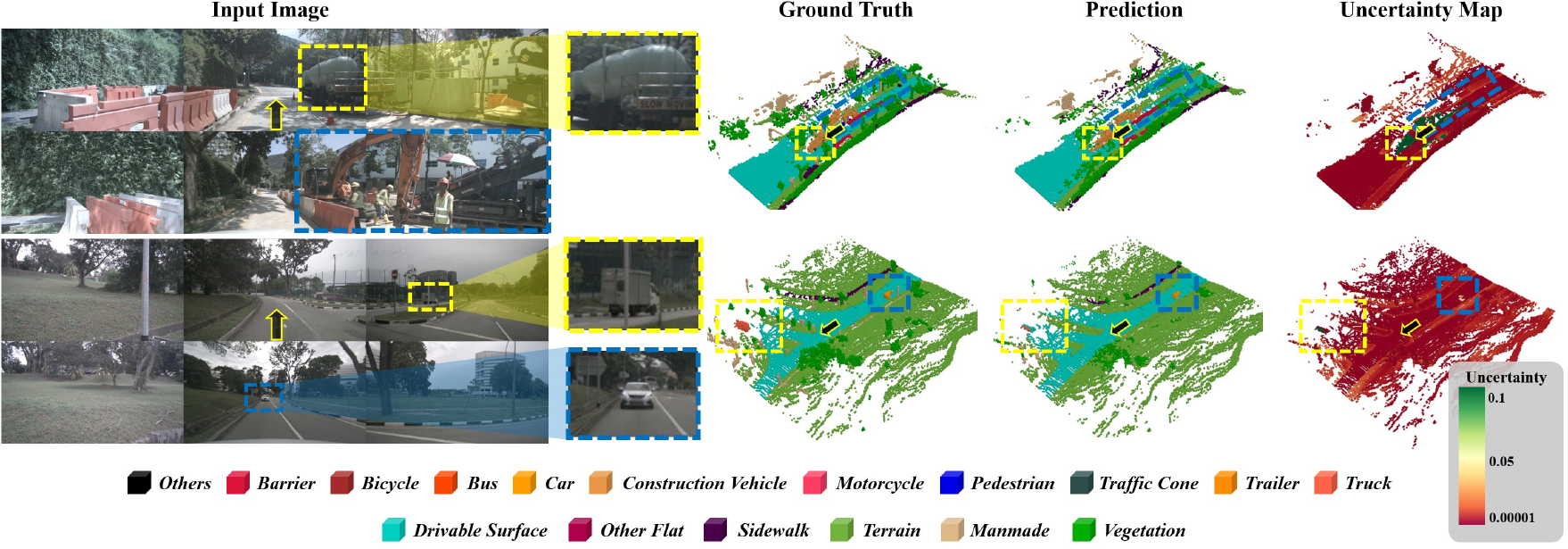}

    \caption{\textbf{Frequency-weighted uncertainty visualization.} Two representative training samples are shown. \textbf{Top:} A high-uncertainty sample with diverse classes, including rare objects. \textbf{Bottom:} A low-uncertainty sample dominated by common classes. Frequency weighting highlights rare-class regions and suppresses dominant backgrounds. Dashed boxes indicate zoomed-in regions.}
    \label{fig:combiend_score}
\end{figure*}
\subsection{Ablation Studies}
\paragraph{Ablation on acquisition criteria}
\Cref{table:ablation_cyclewise_geo_split} shows that the three criteria interact in a non-additive way under the geo-disjoint split. 
Using $U_{\mathrm{fw}}$ alone already provides a strong baseline and consistently improves over Random across cycles.
As illustrated in \Cref{fig:combiend_score}, our frequency-weighted uncertainty concentrates scores on minority-class regions with ambiguous boundaries while suppressing background-dominated areas, making $U_{\mathrm{fw}}$ informative even under geographical shift.
However, adding only one diversity term ($D_{\mathrm{inter}}$ or $D_{\mathrm{intra}}$) does not consistently yield additional gains over $U_{\mathrm{fw}}$, suggesting a trade-off between exploiting hard, rare-class regions and enforcing diversity.

$D_{\mathrm{inter}}$ encourages coverage of different sample-level compositions, which can favor samples that look diverse at the histogram level but contain limited rare foreground content. 
$D_{\mathrm{intra}}$ reduces redundancy within a selected batch, but it may also spread the budget across moderately informative samples when several similar hard cases would be more effective early on. 

When the two diversity metrics are used without the uncertainty term, $(D_{\mathrm{inter}}{+}D_{\mathrm{intra}})$ achieves strong performance and exceeds either component alone. $D_{\mathrm{inter}}$ spreads class distributions across the selected pool, and $D_{\mathrm{intra}}$ removes redundancy within the set based on differences in class composition. Together they form a complementary exploration policy that chooses samples with diverse semantic content while avoiding repetitive sampling.

The best performance is obtained only when all three terms are combined in CAS (Eq.~\eqref{eq:cas_full}), indicating that uncertainty and diversity play complementary roles rather than additive ones. 
Specifically, $U_{\mathrm{fw}}$ drives acquisition toward rare-class errors, $D_{\mathrm{inter}}$ avoids repeatedly sampling already-covered class compositions, and the conditional $D_{\mathrm{intra}}$ removes within-cycle redundancy during greedy selection.

\input{tables/ablation_study_geo_splits}
\input{tables/ablation_top_k_geo_split}

\paragraph{Impact of initial labeled set size}
\Cref{table:ablation_init_cycles} varies the initial labeled seed from 2{,}000 to 8{,}000 samples.
Larger seeds improve the first cycle, since the initial model is trained with broader coverage.
As active learning proceeds, the gap narrows and the results converge.
At Cycle~5, starting from 2{,}000-sample seed, our method reaches 26.62 mIoU with 10k labeled samples, while starting from 8{,}000-sample seed, it reaches 26.66 mIoU with 16k labeled samples.
This indicates that the acquisition policy progressively adds informative and non-redundant samples, so the final performance depends more on the acquired samples
than on the initial seed size.

%%%%%%%%%%%%%%%%%%%%%%%%%%%%%%%%%%%%%%%%%%%%%%%%%%%%%%%%%%%%%%%%%%%%%%%%%%%%%%%%%%%%%%%%%%%%%%%%%%%%%%%%%%%%%%%%%%%%%%%%%%%%%%%%%%%

\section{Conclusion}
We proposed an active learning framework for 3D occupancy prediction that addresses high annotation costs and severe class imbalance in autonomous driving. Our approach integrates frequency-weighted uncertainty with inter-sample and intra-set diversity metrics to prioritize informative samples effectively. Experiments on Occ3D-nuScenes demonstrate substantial improvements over existing methods, particularly for safety-critical minority classes including pedestrians, traffic cones, and bicycles. The complementarity of our metrics, validated through ablation studies, enables significant labeling reduction while maintaining model performance. \newline
\textbf{Limitations}: Our method degrades in night samples, where poor illumination reduces prediction reliability and affects class-distribution estimation. In addition, diversity is measured at the sample-level histogram rather than explicit spatial embeddings, which may not fully capture fine-grained spatial layout similarities. Incorporating geometry-aware or embedding-based diversity into the acquisition score is a promising direction for future work.

%% file: tables/main_results_geo_split.tex
\begin{table*}[t!]
    \centering
    \footnotesize
    \setlength{\tabcolsep}{2.5pt}
    \renewcommand{\arraystretch}{1.12}
    \newcommand{\classfreq}[1]{{~\tiny(\semkitfreq{#1}\%)}} 
    
    \caption{Performance Comparison of Active Learning Methods on the Occ3D-nuScenes validation set~\cite{occ3d},
    constructed using a geographically disjoint split (train/val non-overlapping) following the protocol of Lilja \emph{et al.}~\cite{lilja2024localization}.
    ‘Const. Veh.’ denotes Construction Vehicles, while ‘Drive. Surf.’ represents Drivable Surfaces.}
    
    % \resizebox{\linewidth}{!}{
    \begin{tabular}{l|c|c|ccccccccccccccccc}
    \toprule
    \makecell{\textbf{Cycle} \\ ($N$ samples, \%)}
    & \textbf{AL Method}
    & \textbf{mIoU}
    & \rotatebox{90}{\textcolor{others}{$\blacksquare$} others}
    & \rotatebox{90}{\textcolor{barrier}{$\blacksquare$} barrier}
    & \rotatebox{90}{\textcolor{bicycle}{$\blacksquare$} bicycle}
    & \rotatebox{90}{\textcolor{bus}{$\blacksquare$} bus}
    & \rotatebox{90}{\textcolor{car}{$\blacksquare$} car}
    & \rotatebox{90}{\textcolor{const. veh.}{$\blacksquare$} const. veh.}
    & \rotatebox{90}{\textcolor{motorcycle}{$\blacksquare$} motorcycle}
    & \rotatebox{90}{\textcolor{pedestrian}{$\blacksquare$} pedestrian}
    & \rotatebox{90}{\textcolor{traffic cone}{$\blacksquare$} traffic cone}
    & \rotatebox{90}{\textcolor{trailer}{$\blacksquare$} trailer}
    & \rotatebox{90}{\textcolor{truck}{$\blacksquare$} truck}
    & \rotatebox{90}{\textcolor{drive. surf.}{$\blacksquare$} drive. surf.}
    & \rotatebox{90}{\textcolor{other flat}{$\blacksquare$} other flat}
    & \rotatebox{90}{\textcolor{sidewalk}{$\blacksquare$} sidewalk}
    & \rotatebox{90}{\textcolor{terrain}{$\blacksquare$} terrain}
    & \rotatebox{90}{\textcolor{manmade}{$\blacksquare$} manmade}
    & \rotatebox{90}{\textcolor{vegetation}{$\blacksquare$} vegetation} \\
    \midrule
    
    %% == Cycle 1 == %%
    \makecell[c]{\textbf{Cycle 1} \\ (2,000, 8.5\%)}
    & Random 
    & 18.33 & 0.25 & 30.48 & 0.00 & 23.79 & 37.11 & 0.15 & 0.00 & 6.53 & 0.00 & 11.61 & 23.93 & 68.64 & 8.85 & 30.47 & 31.99 & 21.20 & 16.52 \\
    \midrule
    
    %% == Cycle 2 == %%
    \multirow{4}{*}{\makecell[c]{\textbf{Cycle 2} \\ (4,000, 17.0\%)}}
    & Random 
    & 23.35 & 1.31 & \textbf{36.37} & \textbf{0.92} & 33.18 & 40.03 & 10.43 & 3.95 & \textbf{14.52} & 10.12 & \textbf{26.03} & 28.42 & \textbf{70.54} & 11.82 & 33.03 & 34.60 & 22.31 & \textbf{19.33} \\
    & Entropy 
    & 22.70 & 2.01 & 35.08 & 0.33 & 32.24 & 39.04 & 7.94 & \textbf{7.98} & 13.79 & 7.89 & 23.61 & 27.57 & 70.13 & 10.87 & 32.48 & 34.25 & 22.23 & 18.49 \\
    & Coreset\cite{coreset} 
    & 22.69 & 2.02 & 35.77 & 0.04 & 31.26 & 39.70 & 10.77 & 1.40 & 12.96 & 8.90 & 23.11 & 28.25 & 70.37 & 11.45 & 33.28 & \textbf{34.68} & \textbf{22.55} & 19.20 \\
    & Ours 
    & \textbf{24.17} & \textbf{2.75} & 35.72 & 0.81 & \textbf{34.04} & \textbf{40.15} & \textbf{16.84} & 5.00 & 14.51 & \textbf{12.78} & 25.94 & \textbf{30.21} & 70.32 & \textbf{12.43} & \textbf{33.85} & 34.22 & 22.40 & 18.99 \\
    \midrule
    
    %% == Cycle 3 == %%
    \multirow{4}{*}{\makecell[c]{\textbf{Cycle 3} \\ (6,000, 25.4\%)}} 
    & Random 
    & 24.40 & 3.38 & 35.88 & 0.89 & 32.87 & 40.21 & 13.64 & 8.84 & 14.45 & 12.04 & 27.01 & 28.83 & 71.02 & 12.50 & 34.26 & 35.43 & \textbf{23.20} & 20.30 \\
    & Entropy 
    & 24.54 & 2.63 & 35.24 & 4.84 & 34.36 & 39.72 & 13.80 & \textbf{12.43} & 13.95 & \textbf{13.02} & 24.56 & 27.72 & 70.94 & 12.62 & 33.99 & 34.96 & 23.00 & 19.36 \\
    & Coreset\cite{coreset} 
    & 24.41 & 3.04 & 36.16 & 1.25 & 33.54 & 40.13 & 11.85 & 10.28 & \textbf{15.52} & 11.87 & 25.70 & 29.31 & \textbf{71.24} & 12.82 & 34.35 & \textbf{35.45} & 22.84 & 19.61 \\
    & Ours 
    & \textbf{25.57} & \textbf{3.83} & \textbf{37.07} & \textbf{5.55} & \textbf{34.61} & \textbf{40.58} & \textbf{17.72} & 11.74 & 14.55 & 12.51 & \textbf{28.10} & \textbf{30.69} & 70.85 & \textbf{13.63} & \textbf{34.41} & 35.22 & 23.02 & \textbf{20.59} \\
    \midrule
    
    %% == Cycle 4 == %%
    \multirow{4}{*}{\makecell[c]{\textbf{Cycle 4} \\ (8,000, 34.0\%)}}
    & Random 
    & 25.84 & 4.89 & 36.68 & \textbf{8.88} & \textbf{35.11} & 40.97 & 13.44 & 12.17 & 15.42 & 15.56 & 26.52 & 29.23 & 71.67 & 13.83 & 35.20 & 35.90 & 23.11 & 20.64 \\
    & Entropy 
    & 25.02 & 2.74 & 35.67 & 7.44 & 33.53 & 39.85 & 13.14 & 12.92 & 14.24 & 11.73 & 25.20 & 29.19 & 71.50 & \textbf{14.46} & 35.00 & 35.63 & 23.13 & 19.99 \\
    & Coreset\cite{coreset} 
    & 25.43 & 3.48 & 36.16 & 5.10 & 34.23 & 40.40 & 14.68 & 13.13 & 16.02 & 15.27 & 25.63 & 29.00 & 71.56 & 13.05 & \textbf{35.55} & 35.82 & 23.05 & 20.26 \\
    & Ours 
    & \textbf{26.38} & \textbf{4.99} & \textbf{36.85} & 7.38 & 34.61 & \textbf{40.99} & \textbf{17.49} & \textbf{14.74} & \textbf{16.27} & \textbf{15.59} & \textbf{27.29} & \textbf{30.80} & \textbf{71.92} & 13.72 & 35.52 & \textbf{35.98} & \textbf{23.58} & \textbf{20.78} \\
    \midrule
    
    %% == Cycle 5 == %%
    \multirow{4}{*}{\makecell[c]{\textbf{Cycle 5} \\ (10,000, 42.4\%)}}
    & Random 
    & 25.67 & 4.38 & 36.09 & 5.79 & 34.16 & 40.20 & 15.61 & 12.32 & 15.17 & 15.34 & 26.16 & 29.99 & 71.68 & 13.71 & 35.48 & 35.98 & 23.29 & \textbf{21.02} \\
    & Entropy 
    & 25.91 & 3.45 & 36.00 & 7.74 & 34.83 & 39.68 & 16.08 & \textbf{14.90} & 14.54 & 13.49 & \textbf{27.36} & 29.94 & 71.73 & \textbf{14.76} & 35.60 & \textbf{36.27} & \textbf{23.48} & 20.57 \\
    & Coreset\cite{coreset} 
    & 25.96 & \textbf{4.97} & 36.26 & 6.36 & 35.15 & 40.95 & 15.97 & 13.41 & 16.07 & 15.82 & 25.03 & 29.87 & \textbf{71.96} & 13.93 & 35.54 & 35.89 & 23.24 & 20.84 \\
    & Ours 
    & \textbf{26.62} & 3.73 & \textbf{36.90} & \textbf{11.00} & \textbf{35.29} & \textbf{41.02} & \textbf{18.11} & 14.39 & \textbf{17.09} & \textbf{16.06} & 26.53 & \textbf{30.80} & 71.93 & 13.82 & \textbf{35.77} & 35.89 & 23.28 & 20.88 \\
    \midrule
    
    \makecell[c]{\textbf{Fully Sup.}\\(23,574, 100\%)} & – 
    & 26.83 & 4.47 & 35.58 & 10.31 & 35.12 & 40.81 & 15.18 & 16.72 & 15.79 & 16.93 & 25.87 & 30.14 & 73.42 & 15.31 & 37.45 & 37.38 & 23.74 & 21.89 \\
    \bottomrule
    \end{tabular}
    % }
    \label{table:main_results_geo_split}
\end{table*}

%% file: tables/main_results_semkitti_short.tex
\sisetup{
  detect-all,
  detect-weight = true,
  table-number-alignment = center,
  table-format = 2.2
}

\begin{table}[tb]
  \centering
  \caption{Cycle-wise Semantic Scene Completion (SSC) mIoU on the SemanticKITTI~\cite{behley2019semantickitti} \textbf{validation} set using OccFormer~\cite{zhang2023occformer}.
  Cumulative labeled ratios at each cycle are shown in parentheses.}
  \label{table:main_results_semkitti_short}
  \footnotesize
  \setlength{\tabcolsep}{4pt}
  \renewcommand{\arraystretch}{1.1}

  % change: l -> c (AL Method column centered)
  \begin{tabular}{c|ccccc|c}
    \toprule
    \textbf{AL Method}
      & \makecell[c]{\textbf{Cycle 1}\\(7.8\%)}
      & \makecell[c]{\textbf{Cycle 2}\\(15.6\%)}
      & \makecell[c]{\textbf{Cycle 3}\\(23.4\%)}
      & \makecell[c]{\textbf{Cycle 4}\\(31.2\%)}
      & \makecell[c]{\textbf{Cycle 5}\\(39.0\%)}
      & \makecell[c]{\textbf{Fully}\\\textbf{Sup.}} \\
    \midrule

    Random
      & \multirow{4}{*}{9.66}
      & 10.13 & 10.50 & 10.69 & 11.17
      & \multirow{4}{*}{13.46} \\

    Entropy
      &
      & 10.32 & 10.81 & 11.24 & 11.35
      & \\

    Coreset\cite{coreset} 
      &
      & 10.74 & 11.25 & 11.74 & 11.77
      & \\

    Ours
      &
      & \textbf{10.79} & \textbf{11.82} & \textbf{12.23} & \textbf{12.51}
      & \\

    \bottomrule
  \end{tabular}
\end{table}

%% file: tables/night_rainy_results_geo_split.tex
\definecolor{countgray}{gray}{0.45}
\newcommand{\cnt}[1]{\textcolor{countgray}{\scriptsize #1}}
\newcommand{\hdrgray}[1]{\textcolor{countgray}{#1}}

\begin{table}[t!]
  \centering
  \caption{Cycle-wise mIoU on the Occ3D-nuScenes geographically disjoint validation split, evaluated on the \textbf{Night} and \textbf{Rainy} subsets.
  C2--C5 correspond to Cycle 2--5. Values show mIoU (top) and number of selected samples (\hdrgray{\#samples}, bottom). \\ Best is in \textbf{bold}.}
  \label{tab:night_rainy_results_geo_split}

  \scriptsize
  \setlength{\tabcolsep}{3.0pt}
  \renewcommand{\arraystretch}{1.24}
  \setlength{\aboverulesep}{0.2ex}
  \setlength{\belowrulesep}{0.2ex}
  \setlength{\cmidrulesep}{0.15em}

\begin{tabular*}{\columnwidth}{@{\extracolsep{\fill}}
  >{\centering\arraybackslash}m{1.65cm}@{}|@{}cccc|cccc}
\toprule
\multirow{2}{*}{\textbf{AL Method}} &
\multicolumn{4}{c|}{\textbf{Night} (\textbf{mIoU} / \hdrgray{\#samples})} &
\multicolumn{4}{c}{\textbf{Rainy} (\textbf{mIoU} / \hdrgray{\#samples})} \\
\cline{2-9}
& \textbf{C2} & \textbf{C3} & \textbf{C4} & \textbf{C5}
& \textbf{C2} & \textbf{C3} & \textbf{C4} & \textbf{C5} \\
\midrule

    \multirow{2}{*}{Random}
      & 13.54 & 14.31 & \textbf{14.54} & 14.23
      & 24.96 & 26.33 & \textbf{27.55} & 26.88 \\
      & \cnt{229} & \cnt{214} & \cnt{215} & \cnt{237}
      & \cnt{355} & \cnt{406} & \cnt{402} & \cnt{339} \\
    \cmidrule(l{0pt}r{0pt}){1-9}

    \multirow{2}{*}{Entropy}
      & \textbf{14.74} & \textbf{14.48} & 14.26 & \textbf{14.37}
      & 24.44 & 25.96 & 26.64 & 27.13 \\
      & \cnt{795} & \cnt{525} & \cnt{149} & \cnt{275}
      & \cnt{464} & \cnt{337} & \cnt{374} & \cnt{407} \\
    \cmidrule(l{0pt}r{0pt}){1-9}

    \multirow{2}{*}{Coreset~\cite{coreset}}
      & 13.10 & 13.76 & 13.91 & 14.07
      & 24.65 & 26.38 & 26.66 & 27.41 \\
      & \cnt{164} & \cnt{135} & \cnt{112} & \cnt{190}
      & \cnt{401} & \cnt{412} & \cnt{439} & \cnt{400} \\
    \cmidrule(l{0pt}r{0pt}){1-9}

    \multirow{2}{*}{Ours}
      & 13.42 & 13.57 & 13.78 & 14.04
      & \textbf{26.18} & \textbf{27.24} & 27.32 & \textbf{27.79} \\
      & \cnt{59} & \cnt{95} & \cnt{149} & \cnt{107}
      & \cnt{363} & \cnt{372} & \cnt{448} & \cnt{458} \\
    \bottomrule
  \end{tabular*}
\end{table}

%% file: tables/ablation_study_geo_splits.tex
\begin{table}[t!]
  \centering
  \caption{Ablation study of frequency-weighted uncertainty ($U_{\mathrm{fw}}$),
  inter-sample diversity ($D_{\mathrm{inter}}$), and intra-set diversity ($D_{\mathrm{intra}}$)
  across active learning cycles on the Occ3D-nuScenes geographically disjoint split validation set.}
  \label{table:ablation_cyclewise_geo_split}

  \footnotesize
  \setlength{\tabcolsep}{4.0pt}
  \renewcommand{\arraystretch}{1.20}
  \setlength{\aboverulesep}{0.2ex}
  \setlength{\belowrulesep}{0.2ex}
  \setlength{\cmidrulesep}{0.15em}

  \begin{tabular}{@{}ccc@{\hspace{4pt}}|cccc@{}}
    \toprule
    \multirow{2}{*}{\textbf{$U_{\mathrm{fw}}$}} &
    \multirow{2}{*}{\textbf{$D_{\mathrm{inter}}$}} &
    \multirow{2}{*}{\textbf{$D_{\mathrm{intra}}$}} &
    \multicolumn{4}{c}{\textbf{mIoU (\%)}} \\
     \cline{4-7}
     &  &  & \textbf{Cycle 2} & \textbf{Cycle 3} & \textbf{Cycle 4} & \textbf{Cycle 5} \\
    \midrule

     &  &  & 23.35 & 24.40 & 25.84 & 25.67 \\
    \midrule

    \checkmark &  &  & 23.55 & 24.93 & 26.00 & 26.15 \\
     & \checkmark &  & 23.54 & 24.65 & 25.99 & 25.96 \\
     &  & \checkmark & 23.42 & 24.70 & 25.96 & 26.08 \\
    \midrule

    \checkmark & \checkmark &  & 23.48 & 24.82 & 25.98 & 26.10 \\
    \checkmark &  & \checkmark & 23.51 & 24.95 & 25.90 & 26.07 \\
     & \checkmark & \checkmark & 23.75 & 24.29 & 26.18 & 26.35 \\
    \midrule

    \checkmark & \checkmark & \checkmark &
    \textbf{24.17} & \textbf{25.57} & \textbf{26.38} & \textbf{26.62} \\
    \bottomrule
  \end{tabular}
\end{table}

%% file: tables/ablation_top_k_geo_split.tex
\setlength{\tabcolsep}{2.4pt}
\renewcommand{\arraystretch}{1.24}
\setlength{\aboverulesep}{0.2ex}
\setlength{\belowrulesep}{0.2ex}
\setlength{\cmidrulesep}{0.15em}

\newcommand{\boldnum}[1]{{\bfseries #1}}

\newcommand{\cycgap}{\hspace{1.45em}}
\newcommand{\numgap}{\hspace{0.18em}}

\newcommand{\kmark}[1]{%
  \makebox[0pt][l]{\numgap\raisebox{0.85ex}{\scriptsize\textcolor{gray}{#1}}}%
}

\newcommand{\miouhead}{%
  \textbf{mIoU}\kern0.15em%
  \textsuperscript{\raisebox{0.15ex}{\scriptsize\textcolor{gray}{$N_L$}}}%
}

\begin{table}[t!]
  \caption{
  Cycle-wise mIoU (\%) for different initial labeled set sizes on the
  \textbf{Occ3D-nuScenes geographically disjoint split} validation set (FlashOCC).
  Gray superscripts denote cumulative labeled samples ($\textcolor{gray}{N_L}$), where $\textcolor{gray}{k}$ denotes thousands.
  }
  \label{table:ablation_init_cycles}
  \centering

  \begin{tabular}{@{}c|l@{\cycgap}l@{\cycgap}l@{\cycgap}l@{\cycgap}l@{}}
    \toprule
    \multirow{2}{*}{\textbf{Init. Size}} &
    \multicolumn{5}{c}{\miouhead} \\
    \cline{2-6}
     & \textbf{Cycle 1} & \textbf{Cycle 2} & \textbf{Cycle 3} & \textbf{Cycle 4} & \textbf{Cycle 5} \\
    \midrule

    2{,}000
      & 18.33\kmark{2k}
      & 24.17\kmark{4k}
      & 25.57\kmark{6k}
      & 26.38\kmark{8k}
      & 26.62\kmark{10k} \\

    4{,}000
      & 20.12\kmark{4k}
      & 23.77\kmark{6k}
      & 25.23\kmark{8k}
      & 26.19\kmark{10k}
      & 26.63\kmark{12k} \\

    6{,}000
      & 22.94\kmark{6k}
      & 25.22\kmark{8k}
      & 25.63\kmark{10k}
      & 26.18\kmark{12k}
      & 26.60\kmark{14k} \\

    8{,}000
      & \boldnum{24.36}\kmark{8k}
      & \boldnum{25.81}\kmark{10k}
      & \boldnum{26.12}\kmark{12k}
      & \boldnum{26.40}\kmark{14k}
      & \boldnum{26.66}\kmark{16k} \\

    \midrule
    \textbf{Fully sup.} & \multicolumn{5}{c}{26.83\kmark{23.5k}} \\
    \bottomrule
  \end{tabular}
\end{table}